\documentclass[pdflatex,sn-mathphys-num]{sn-jnl}

\usepackage{multirow}%
\usepackage{amsmath,amssymb}%
\usepackage{xcolor}%
\usepackage{booktabs}%

\def\showrevision{}
\ifdefined\showrevision

\usepackage{xcolor}
\usepackage[normalem]{ulem}
\usepackage{marginnote}

\usepackage{xspace}

\newcommand{\revref}[2]{%
\marginnote{$R_{#1}C_{#2}$}
}
\DeclareRobustCommand{\robustrevref}[2]{%
\revref{#1}{#2}
}

\newcommand{\reverserevref}[2]{%
\normalmarginpar
\ifodd\value{page}
  \marginnote{\hspace*{1cm}$R_{#1}C_{#2}$}
\else
  \marginnote{\hspace*{-1cm}$R_{#1}C_{#2}$}
\fi
\reversemarginpar
}

\newcommand{\revdel}[1]{%
{\color{darkgray}\sout{#1}}\xspace
}

\else

\usepackage{xspace}

\newcommand{\marginnote}[1]{\ignorespaces}
\newcommand{\revref}[2]{\ignorespaces}
\newcommand{\robustrevref}[2]{\ignorespaces}

\newcommand{\revdel}[1]{\ignorespaces}

\fi

\DeclareMathOperator*{\argmin}{arg\,min}

\usepackage{xspace}
\newcommand{\roboreg}{\emph{roboreg}\xspace}

\newcommand{\figref}[1]{\figurename~\ref{#1}}
\newcommand{\tabref}[1]{\tablename~\ref{#1}}

\begin{document}

\title[Streamlining stereo differentiable rendering]{Streamlining stereo differentiable rendering for marker-free real-time tracking of surgical robots}


\author[1]{\fnm{Yanghe} \sur{Hao}}
\equalcont{These authors contributed equally to this work.}

\author[1]{\fnm{Martin} \sur{Huber}}
\equalcont{These authors contributed equally to this work.}

\author[1]{\fnm{Christos} \sur{Bergeles}}

\author[1]{\fnm{Tom} \sur{Vercauteren}}

\affil[1]{\orgdiv{School of Biomedical Engineering \& Imaging Sciences}, \orgname{King's College London}, \orgaddress{\city{London}, \country{United Kingdom}}}


\abstract{
\textbf{Purpose:} We evaluate stereo differentiable rendering based pose estimation for marker-free real-time surgical robots tracking, mitigating occlusion-prone marker-based tracking in cluttered surgical environments, potentially improving safety, reducing setup times, and enabling intelligent multi-robot interaction.

\textbf{Methods:} 
This work extends the differentiable rendering based markerless robot pose estimation framework \roboreg for online real-time dynamic tracking in two ways.
(i) Sequential optimisation propagates pose estimates across consecutive frames, with motion-adaptive hyperparameter tuning balancing convergence and precision during estimation.  
(ii) 
Integrate CUDA stream parallelisation 
for segmentation and the optimisation steps 
and combines it with CUDA-graph accelerated segmentation. 
We collect 38 displacement video sequence datasets 
with unobstructed robot and 5 occluded-robot dataset
with static start/end ground-truth pose calibrations and dynamic marker-based reference tracking in between for accuracy evaluation under different scenarios.

\textbf{Results:} Real-time localisation at 30 fps for 1080p video sequence is achieved, accelerating from 14 fps in the vanilla \roboreg, thereby matching the camera frame rate. 
Near-1cm
accuracy is demonstrated, with 1.7 cm translational and 0.6° rotational error against static ground-truth pose calibration; and with 
1.2 cm average 3D error across 
27,460 frames against a marker-based reference standard (1.53 cm in over 1,242 frames in occlusion evaluation).
Our method outperforms FoundationPose by 11\% (63\% in occlusion dataset) in dynamic estimation and 250\% in static estimation, while achieving $6 \times$ faster inference.

\textbf{Conclusion:} We demonstrate real-time high-resolution marker-free tracking of surgical robots through stereo differentiable rendering. Localisation accuracy performed on par with marker-based approaches and improved upon foundational baselines.}




\maketitle

\section{Introduction}\label{sec1}
Modular surgical robots with independently positioned components enable flexible deployment in space-constrained operating rooms. However, current robots' limited spatial awareness has hindered their deployment, with collisions among arms, equipment, and staff disrupting workflows and necessitating system recovery~\cite{huber2025localising}.
Overcoming this limitation would enable automated multi-robot docking with smart positioning guidance, reducing theatre setup time and anaesthesia exposure. Realising these benefits
requires continuous camera-to-robot pose estimation as components may be repositioned for patient-specific configurations. 
The challenge is exacerbated by sterile draping, visual clutter, and frequent occlusions in surgical environments.

Continuous camera-to-robot pose estimation involves extracting 6D spatial information, including 3D translation and rotation relative to a camera coordinate system~\cite{10655554}. Traditional correspondence-based methods~\cite{10655554} establish 2D-3D correspondences via PnP algorithms~\cite{lu2023markerlesscameratorobotposeestimation}, including fiducial-based approaches~\cite{5979561} that use physical markers. However, these methods assume static camera-robot relationships and fail when base positions change during operation~\cite{Chen_2023}.

Deep learning has advanced pose estimation through both correspondence-based methods, such as keypoint detection~\cite{lee2020cameratorobotposeestimationsingle}, and rendering-based approaches~\cite{Chen_2023}. While keypoint methods offer computational efficiency, they are vulnerable to occlusions. Rendering-based methods exploit the full robot geometry through dense correspondence matching for superior accuracy and robustness; however, most work focuses on offline estimation~\cite{huber2025localising} with no evidence of online continuous pose tracking.

This work extends \roboreg~\cite{huber2025localising}, a differentiable rendering pipeline that iteratively aligns segmented robot masks with rendered CAD model projections to optimise camera extrinsics for each frame, exploring online markerless camera-to-robot pose estimation during continuous motion. Frame-by-frame initialisation, inspired by FoundationPose~\cite{10655554}, ensures temporal coherence while compensating for weak performance against abrupt inter-frame movements~\cite{10655554} through a gradient field at segmenting mask boundaries.
Adaptation of parallelised CUDA multi-streaming~\cite{pytorch2024cuda} enables asynchronous GPU operations for real-time performance.

Our contributions include: (i) the deployment of a
marker-free stereo differentiable rendering and segmentation pipeline 
for online surgical robot localisation, (ii) the collection of a dynamic surgical robot displacement dataset in this novel research domain
under varying environmental conditions, including an unobstructed robot and a robot surrounded by surgical instruments while partially occluded by the operator, (iii) 
Achieves real-time pose estimation by parallelising operations via CUDA multi-stream execution within the existing render-based \roboreg framework, preserving the original white-box architecture without fundamental algorithmic modifications,(iv) Experimentally benchmarking our method against the state-of-the-art FoundationPose baseline (RGB-D based) using AprilTag-derived ground truth as the reference for 6D robot pose estimation in the RGB domain.

\section{Related work}\label{sec:relatedwork}
Traditional marker-based pose estimation relies on fiducial markers~\cite{5979561} placed at known locations along the robot's kinematic chain. These methods establish 2D-3D correspondences through marker detection and solve camera-to-robot transformations via PnP optimisation~\cite{Horaud_1995}. Despite their maturity, they require physical modification of robot structures, manual configuration across multiple joint poses, and are susceptible to marker occlusion or detachment~\cite{lee2020cameratorobotposeestimationsingle}.

Early markerless methods exploited depth cameras for dense 3D measurements~\cite{BMVC2015_181}. \citet{6907311} introduced pixel-wise classification of depth images to robot parts using voting-based pose inference. \citet{7487185} employed Random Forest regressors for direct depth-to-joint-angle mapping, demonstrating the effectiveness of learning-based approaches for markerless pose estimation.

Deep learning enabled direct RGB-based pose estimation through keypoint detection methods. These approaches detect predefined robot keypoints (e.g., joint locations) in image space and register them with 3D positions derived from forward kinematics, camera intrinsics and joint states, establishing 2D-3D correspondences for PnP-based pose estimation. DREAM~\cite{lee2020cameratorobotposeestimationsingle}, trained with synthetic data and domain randomisation, predicts 2D joint locations from single RGB frames. However, keypoint methods remain vulnerable to occlusion, motion blur, and environmental variations that compromise detection reliability, particularly for rotational accuracy~\cite{Chen_2023}.

To leverage physical correspondence between the robot geometry and robot poses, point cloud-based methods, such as the iterative closest point (ICP) based algorithm Hydra~\cite{huber2025hydra}, using readily available depth data and various random robot configuration observations, register observed point clouds to the robot's CAD model by minimising a point-to-plane error through iterative refinement on a Lie algebra.

Render-and-compare approaches also leverage CAD models for dense shape matching~\cite{labbé2021singleviewrobotposejoint}, iteratively minimising discrepancies between rendered projections and observed silhouettes through gradient-based optimisation. EasyHeC~\cite{Chen_2023} embeds robot meshes into monocular differentiable pipelines for markerless hand-eye calibration, achieving superior accuracy through full-silhouette optimisation rather than sparse keypoint matching
The recent \roboreg extends the differentiable rendering paradigm to stereo for improved distance estimates, even under few robot calibration configurations~\cite{huber2025localising} 
and under draping.
However, iterative optimisation incurs computational costs, and no prior work has deployed it for online pose estimation. 
State-of-the-art methods aim for 30~Hz inference to match standard RGB camera acquisition rates~\cite{lu2023markerlesscameratorobotposeestimation}, which is essential for rapid response to visual input, particularly in surgical environments where spatial constraints and scene complexity pose additional challenges during preoperative robot deployment.

Hybrid frameworks combine keypoint detection with rendering refinement. CtRNet~\cite{lu2023markerlesscameratorobotposeestimation} initialises differentiable rendering from keypoint predictions, reducing sim-to-real gaps through combined supervision. While it achieving 6.3 cm pose estimation accuracy, outperforming keypoint-based (14.1 cm) and rendering-based (7.4 cm) methods in the Baxter dataset~\cite{lu2023markerlesscameratorobotposeestimation},
performance remains constrained by the quality of the keypoint detector, with reliance on deep networks reduces interpretability.

Foundation models, such as FoundationPose~\cite{10655554}, achieve zero-shot generalisation in 6D object pose tracking via neural implicit representations, with temporal coherence established by propagating pose across frames. However, tracking performance degrades under rapid inter-frame movements. 

Despite the advances, the existing markerless approach still suffers from a trade-off between robustness and processing efficiency. While efficient methods, such as keypoints, have suffered from insufficient robot pose interpretation due to domain gaps, the slow inference speed bottlenecks the more robust rendering methods. Critically, no existing framework achieves real-time operation at runtime while mainly focusing on static calibration.

\section{Method}\label{sec3}

\paragraph{Real-time robot pose tracking}\label{subsec2}

\begin{figure}[tb]
\centering
\includegraphics[width=0.8\textwidth]{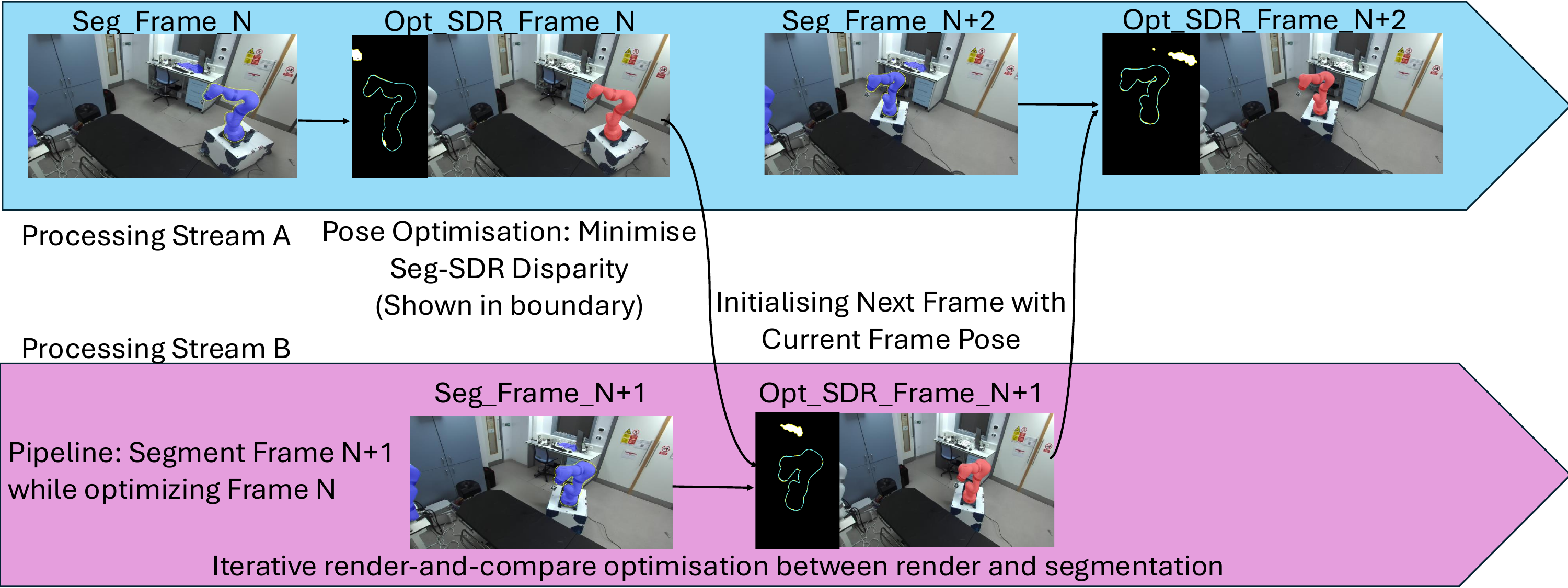}
\caption{Multi-stream pipeline for dynamic camera-to-robot pose estimation~\cite{huber2025localising, pytorch2024cuda}.}
\label{fig:pipeline}
\end{figure}
The proposed stereo-differentiable rendering-based real-time tracking approach is summarised in \figref{fig:pipeline}. Therein, the rendering and segmentation steps are executed concurrently. Details are provided below.

\subparagraph{Stereo-differentiable rendering-based tracking}\label{subsec1}
Stereo-differentiable rendering facilitates finding the pose $\boldsymbol{\Theta}_{l}\in \text{SE}(3)$ of a CAD model (the robot) as observed by a stereo camera (here parametrised in the left camera frame) through gradient-based optimisation:
\begin{equation}
    \argmin_{\boldsymbol{\Theta}_l} f\left(\mathbf{R}_l(\boldsymbol{\Theta}_l), \mathbf{S}_l) + f(\mathbf{R}_r(\boldsymbol{\Theta}_l), \mathbf{S}_r\right)
    \label{eq:objective}
\end{equation}
where $\mathbf{R}_{l/r}$ are respectively left/right silhouette renders, and $\mathbf{S}_{l/r}$ are left/right image segmentations of the robot. The objective function $f$ is further described below. We exploit this iterative gradient-based optimisation approach for online pose tracking. Here, we implement stereo-differentiable rendering using the \emph{roboreg} library\footnote{Available open-source at \url{https://github.com/lbr-stack/roboreg}}.
render and optimise using both left- and right-camera views.
Each frame's optimisation is initialised from the converged pose of the previous frame to exploit temporal continuity. As robot joints remain fixed throughout the experiment, the configuration is initialised once at the beginning, reducing per-iteration computation to a single camera transformation and enabling more efficient optimisation.

\subparagraph{Camera pose initialisation} 
Iterative pose tracking demands an initial camera pose. To overcome the initialisation problem at the start of the process,
we initialise camera poses using the Hydra algorithm~\cite{huber2025hydra}, given readily available depth data, then refine the obtained camera-to-robot extrinsic parameters $\boldsymbol{\Theta}_l$ through stereo-differentiable rendering as above. 
In the proposed dynamic pose tracking, the result of the pose estimation from the previous frame is used as initialisation for pose optimisation in the current frame.
The hand-eye calibration performed at 
the end of the motion sequence will also serve as the static ground truth for robot pose estimation.

\subparagraph{Objective function} 
The original \emph{roboreg} stereo-differentiable rendering pipeline extends the cost function \eqref{eq:objective}, incorporating robot segmentations with an exponentially decaying Euclidean distance-based function to increase the convergence basin. Computing the Euclidean distance transform (EDT), however, is computationally expensive. Given the above robust initialisation and sufficiently fast online pose updates, as facilitated by concurrent optimisation below, we here assume well-overlapping rendered silhouettes and segmentations. Under these relaxed conditions, we forgo the EDT 
as standard pixel-wise losses are able to provide adequate gradients from the assumption of well-overlapping
and directly use the objective function in~\eqref{eq:objective}.
To improve robustness to errors in the segmentation masks, we however switch from the original soft Dice, whereby false positives localised around the robot generate adversarial gradient signals that compromise pose optimisation integrity,
to a Tversky loss~\cite{salehi2017tversky} expressed for the left image as:  
\begin{equation}
f_l = 1 - \frac{\|\mathbf{M}_l \odot \mathbf{S}_l\|_1}{\|\mathbf{M}_l \odot \mathbf{S}_l\|_1 + \alpha \|\mathbf{M}_l \odot (1-\mathbf{S}_l)\|_1 + \beta \|(1-\mathbf{M}_l) \odot \mathbf{S}_l\|_1 + \epsilon}
\end{equation}
where $\mathbf{S}_l$ are segmentation probabilities, $\mathbf{M}_l$ is the rendered silhouette, $\odot$ the Hadamard product, $\epsilon=1\mathrm{e}{-6}$ a numerical constant, $\alpha$ and $\beta$ are coefficients for controlling the false positive and false negative penalty in range [0, 1]
, respectively, which penalise the false positives in the segmentation by tuning the coefficients, and the total loss comprises left and right terms in this stereo formulation: $f = f_l + f_r$.

\subparagraph{Concurrent optimisation} 
To maximise compute utilisation and thus accelerate pose tracking, we suggest concurrent execution across consecutive frames as illustrated in \figref{fig:pipeline}, here implemented via CUDA streams~\cite{pytorch2024cuda}. 
Through concurrent CUDA streams, each sequentially executing the segmentation and differentiable-rendering steps, multiple frames are processed simultaneously. 
Our implementations have utilised the CUDA stream parallelisation concepts  with two streams, designed as  \texttt{stream\_current} and \texttt{stream\_next}, and are managed in a ping-pong pattern. The streams are swapped after each frame completes. At the pipeline level, when frame $N$ finishes segmentation on \texttt{stream\_current}, segmentation immediately launches for frame $N+1$ on \texttt{stream\_next} while the current frame is optimising in parallel, overlapping operations across consecutive frames. Event-based synchronisation ensures that frame $N+1$ optimisation starts only after $N+1$ segmentation and $N$ optimisation are complete, to avoid race conditions. After frame $N$ completes all processing, streams swap roles: \texttt{stream\_current} becomes \texttt{stream\_next} and vice versa, enabling continuous reuse of stream resources throughout the sequence. 
Sequential execution time per frame is
$t_{\text{sequential}} = t_{\text{seg}} + t_{\text{opt}}$.
Multi-stream pipelining reduces per-frame time to
$t_{\text{pipeline}} = \max(t_{\text{seg}}, t_{\text{opt}})$.
The effective frame rate approaches $1/t_{\text{pipeline}}$, reducing processing overhead from sequential stages to a single bottleneck stage with a theoretical speedup of
$\frac{t_{\text{seg}} + t_{\text{opt}}}{\max(t_{\text{seg}}, t_{\text{opt}})}$.
This maximises GPU utilisation by overlapping compute-intensive operations, achieving real-time performance while maintaining per-frame accuracy.

\subparagraph{Motion-aware adaptive optimisation} 
Multi-stream pipelining achieves parallelisation but cannot eliminate the accuracy-latency trade-off. Aggressive gradient steps risk poor convergence, while conservative gradient steps can ensure quality at the cost of increased computational requirement. We propose motion-aware adaptive optimisation that dynamically adjusts optimiser hyperparameters based on inter-frame motion characteristics to achieve context-dependent balancing of accuracy and efficiency across diverse motion conditions.  
For each frame N, we dilate the initialised rendering mask to define a Region of Interest (ROI) and crop the segmentation mask accordingly. We then compare segmentations between consecutive frames N-1 and N (with ROIs similarly defined) for multi-factor motion classification based on three geometric properties: (i)~soft Intersection over Union (IoU) detects overall displacement; 
(ii)~centroid displacement of the segmentation probability mask within the ROI quantifies translational motion; 
and (iii)~angular difference from probability-weighted image moments captures rotation. 
The normalised centriod displacement is computed as $d = \frac{1}{\sqrt{W^2 + H^2}}\sqrt{\left(\frac{\sum_{x,y} x \cdot M^t}{\sum_{x,y} M^t} - \frac{\sum_{x,y} x \cdot M^{t-1}}{\sum_{x,y} M^{t-1}}\right)^2 + \left(\frac{\sum_{x,y} y \cdot M^t}{\sum_{x,y} M^t} - \frac{\sum_{x,y} y \cdot M^{t-1}}{\sum_{x,y} M^{t-1}}\right)^2}$ where $M^t$ and $M^{t-1}$ denote the segmentation probability masks at frames $t$ and $t-1$ respectively, $x$ and $y$ are the global image coordinates of segmentation pixels within the cropped ROI, and $W$ and $H$ are the image width and height used for normalisation.

\paragraph{Implementation details}\label{subsec4}
We utilise the \emph{roboseg-v0-large} segmentation model~\cite{huber2025localising} on images of $1024 \times 576$, which is the highest resolution possible with computation constraints. Obtained segmentation probability maps are then upsampled back to 1080p for differentiable rendering.

Segmentation inference is accelerated by tracing operations through CUDA graphs~\cite{pytorch2024cuda}, recording the computational graph once and replaying it for subsequent host-device transfer overhead reduction.
The optimisation of \eqref{eq:objective} employs AdamW~\cite{loshchilov2017adamw} with motion-aware adaptive hyperparameters. Motion detection is performed with an ROI padded with 5 pixels around the stale rendering mask, computing soft IoU, centroid displacement, and orientation changes between consecutive frames within the ROI, classifying motion into three regimes with corresponding parameter adjustments.
For stable tracking (IoU $\geq 0.995$), conservative learning rate $5 \times 10^{-4}$ with momentum $(\beta_1, \beta_2) = (0.95, 0.999)$ ensures smooth convergence. Rotation detection (low IoU with angular change $> 0.1\deg$) triggers increased learning rate $2.75 \times 10^{-2}$ with reduced momentum $(0.9, 0.95)$ for rapid pose correction. Translation (low IoU with normalised centroid displacement $> 0.001 m$) employs the same aggressive learning rate with slightly higher second momentum $(0.9, 0.98)$. All regimes share weight decay $w = 1 \times 10^{-4}$, maximum iterations set to $N_{\text{iter}} = 5$ and ablation study $N_{\text{iter}} = 10$, and gradient clipping 0.1.

\section{Experiment and results}\label{sec4}
Due to the absence of publicly available datasets for dynamic robot pose estimation, we established a benchmarking dataset, as further detailed below.

\subparagraph{Experimental setup} 
We utilised a KUKA LBR Med 7 platform~\cite{Huber2024} on a moveable base for displacement experiments. The robot was equipped with an AprilTag on the end-effector to establish robust ground-truth tracking. A Stereolabs ZED 2i stereo camera (1920$\times$1080 resolution; $30\,\text{fps}$) was mounted on a tripod for tracking. Inference was performed on a machine running Ubuntu 24.04 with AMD Ryzen 9 5900X processor and NVIDIA RTX 5090 GPU. 

\begin{figure}[tb]
\centering
\includegraphics[width=1\textwidth]{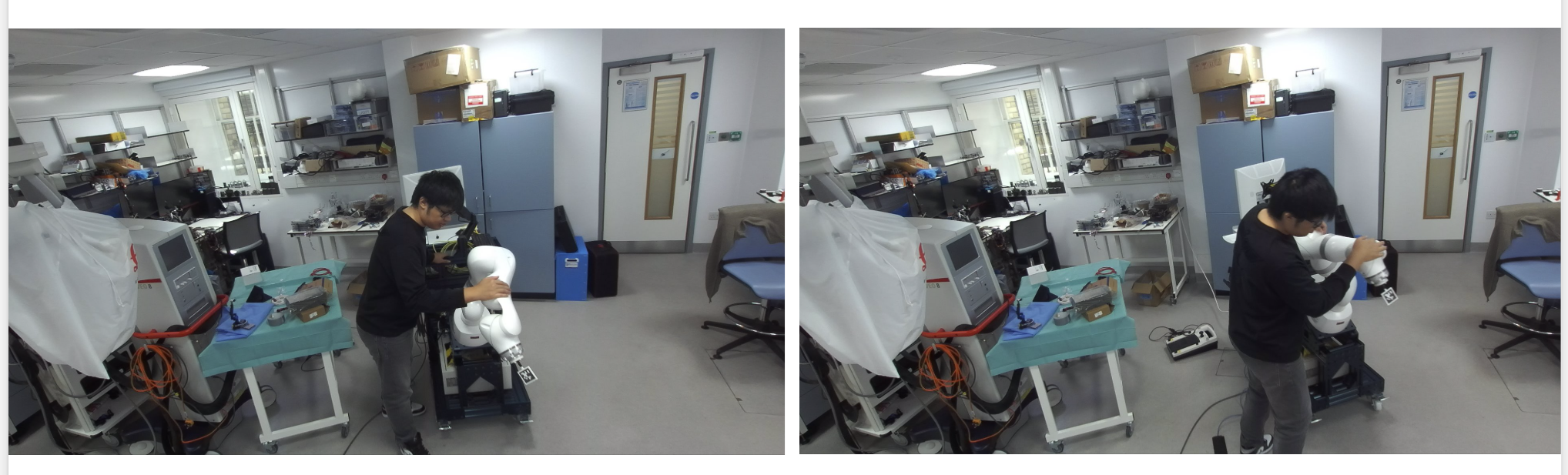}
\caption{
The example video frame obtained in occlusion displacement sequence 0\_1 (left, mild occlusion), 3\_4 (right, heavy occlusion), viewed from the left camera with the operator moving the robot while the body is partially blocking the robot at different level\label{fig:pos_occlude}}
\end{figure}

\subparagraph{Benchmarking data collection} We collected a total of 19 displacement sequences (9 robot displacements, 10 camera displacements). These 19 sequences were played both forward and backward. Our dataset is made openly accessible to serve future research\footnote{Dataset available in open-access at \url{https://doi.org/10.18742/32129134}}, the video source are compressed with H.264 encoder under quasi-lossless settings, totalling 38 investigated sequences.
 Each video sequence is captured at 10 Hz from the ROS2 bag capture, with visual inspection to confirm realism. Each sequence comprises large-scale random translations, rotations, and combinations, producing varied motion speeds and displacements. Rotations alter the robot's visible silhouette and introduce self-occlusions, testing robustness to appearance changes. The capture procedure follows a calibrate–move–calibrate pattern: starting poses initialise online tracking, final poses provide localisation benchmarks and initialisation for subsequent sequences, and reversal enables backward playback evaluation. 
Mid-displacement ground-truth localisation was obtained via the mounted AprilTag, forming the second mode of evaluation in this study.
For a more realistic evaluation, an additional set of five displacement sequences is captured in a more complex environment.
There, the robot is partially occluded by the operator during motion, and additional equipment appears in the footage, including extra medical tables and a large Leica surgical microscope.
This increases visual complexity as shown in \figref{fig:pos_occlude} with different levels of occlusion by the operator. The occlusion patterns comprise two regimes: mild occlusion (\figref{fig:pos_occlude}, left), in which the operator's hand rests on a single joint (Sequences 0\_1 and 2\_3); and severe occlusion (\figref{fig:pos_occlude}, right), in which the operator's body occasionally blocks a substantial portion of the robot, where multiple joints and links are not seen by the camera during the motion (Sequences 1\_2, 3\_4, and 4\_5). These sequences follow similar movement patterns to those in the prior datasets and are evaluated using the same AprilTag-based ground-truth methodology.

\subparagraph{Comparative baseline}
For comparison with a state-of-the-art approach, we rely on FoundationPose~\cite{10655554} model-based tracking on the same video datasets.
Existing state-of-the-art methods (DREAM, EasyHec, CtRNet) are constrained to specific robot platforms and are incompatible with our LBR Med 7, precluding direct comparison.  FoundationPose was selected as the baseline for its platform-agnostic design, requiring only CAD meshes, 
RGB-D input, and an initial object segmentation mask without task-specific retraining,
while supporting both static pose initialisation with meshes and temporal frame-to-frame tracking during robot motion.
The initial target segmentation is provided by the same segmentation model used in our approach.
This establishes a robot-agnostic baseline reflecting realistic deployment where hardware selection is independent of algorithm compatibility. The robot mesh is initialised at 640×480 resolution (upsampled to 1080p) on an RTX 4060 GPU, using depth maps from FoundationStereo~\cite{wen2025foundationstereo} to improve accuracy and reduce overhead compared to ZED depth computation. Temporal pose estimation proceeds in tracking mode.

\subparagraph{Evaluation metrics} The experimental evaluation encompasses three components: 
(i) computational efficiency analysis comparing sequential versus multi-stream processing, 
(ii) 6D pose estimation accuracy through comparison with final estimation of dynamic pose estimation to static \roboreg calibration ground truth, (iii) dynamic pose accuracy benchmarking against AprilTag-based reference standard during motion sequences.

The experimental evaluation assesses computational efficiency and pose estimation accuracy. Computational performance compares overlapping FPS from pipelined multi-stream inference against a single-thread FPS baseline, with a speedup ratio quantifying architectural efficiency.

For static evaluation, the final dynamic pose estimates are compared with initial calibration checkpoints using 6D pose metrics for translation and rotation to quantify pose estimation accuracy,
against the baseline performance of FoundationPose tracking of the final pose. 
For dynamic evaluation, AprilTag pose is estimated via Stereo PnP, jointly optimising 6-DoF pose by minimising reprojection error across both views through trust-region least squares, initialised with monocular PnP. Corner detection uses Pupil-AprilTag~\cite{5979561}, validated against 200 manually labelled frames. Positional error is computed between the Stereo PnP estimate and expected tag pose derived from estimated extrinsics, forward kinematics, and end-effector-to-tag transformation
across all frames in the displacement dataset.
Signed error analysis (forward kinematics minus tag-derived positions) quantifies systematic biases, where positive and negative errors indicate overestimation and underestimation, respectively. Metrics include 3D Euclidean distance, component-wise mean absolute error (MAE) along Z: depth; X/Y: horizontal/vertical in camera coordinates
and statistical distributions (mean, median, standard deviation) characterising axis-specific bias.
Ablation studies validate each module by varying key parameters and at higher iterations: resolution variants (1080p vs. segmentation-matched downsampling), 
fixed hyperparameter configurations (base learning rate $5 \times 10^{-3}$) against adaptive settings, and a fixed aggressive mode (learning rate $2.75 \times 10^{-2}$) quantifying error accumulation under sustained high learning rates.
The occlusion study will set the optimiser to adaptive mode, with the iteration count set to 5.
\paragraph{Inference efficiency analysis}\label{sec:efficiencyanalysis}
The system achieves 34.0~FPS at 1080p across all 19 sequences, with parallelisation accelerating processing from 17.2~FPS (sequential) to 34.0~FPS—a $1.97\times$ speedup exceeding the ZED 2i's 30~FPS maximum. Component analysis reveals 33.3~ms segmentation latency and 259.9~iter/s optimisation rate, nearly doubling the original \roboreg implementation with per-frame mesh updates and $6\times$ faster than FoundationPose on the same dataset at a downsampled resolution, 
while also outperforming the 30-second inference time per pose in EasyHec and 1 Hz loop rate of differentiable rendering approach reported in CtrNet.

\paragraph{Pose estimation evaluation against static calibrations}\label{subsec3}
As configured in the dataset, static calibration using the multi-pose static calibration baseline is performed at the beginning and end of the robot's movement, serving as a ground-truth checkpoint for estimating the model-derived poses after motion. 
Evaluation spans 16 sequences (32 trajectories, forward and reverse), excluding three sequences corrupted due to connection issue during capture. Static ground truth is derived from multi-pose Hydra calibration~\cite{huber2025hydra}, refined with stereo \roboreg at final positions of each trajectory, and compared against final pose estimates after iterating over all frames. Results are summarised in \tabref{tab:comparison}.

\begin{figure}[tb]
\centering
\includegraphics[width=1\textwidth]{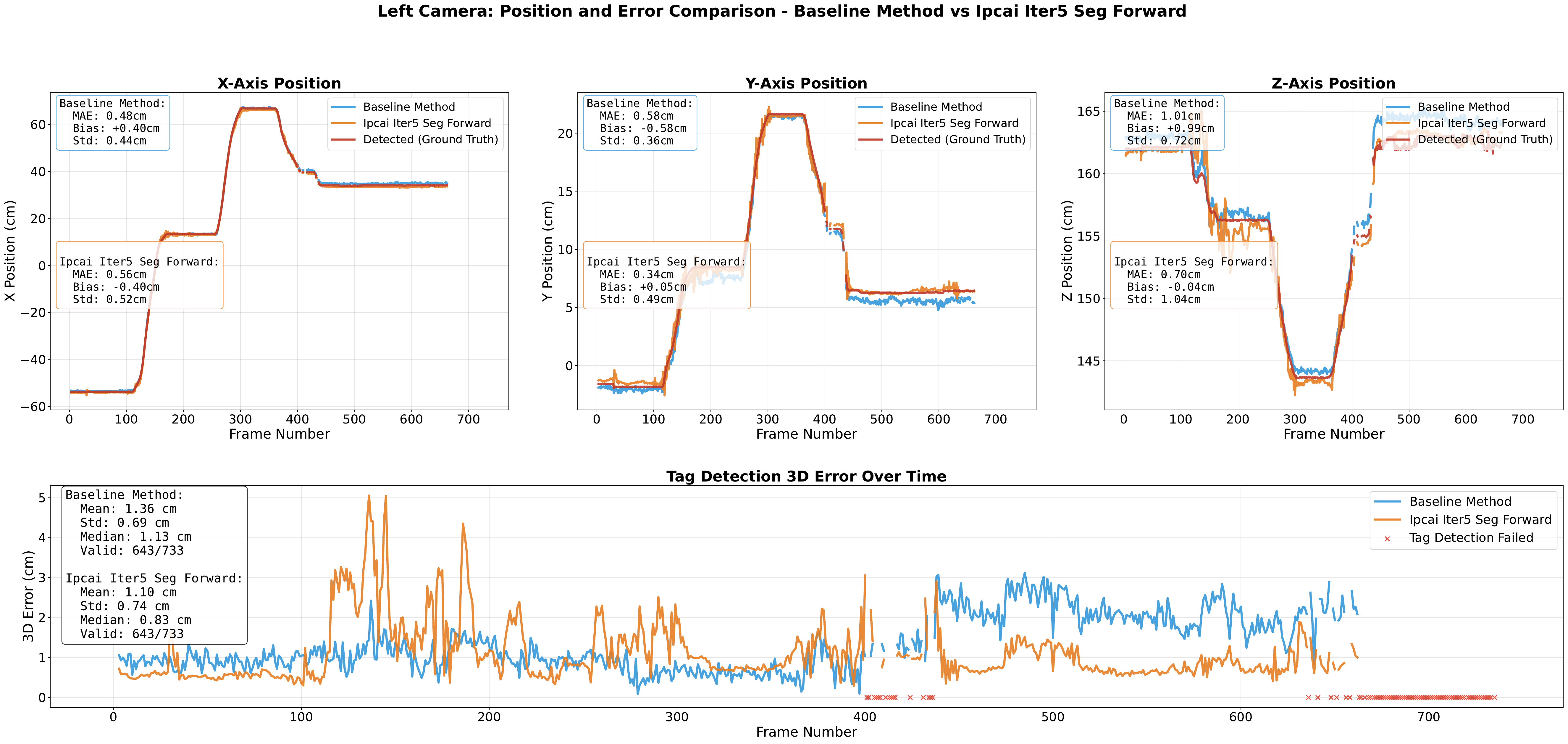}
\caption{Detailed positions for robot motion from position 11 to position 12 of our dataset in x,y,z direction with component-wise error comparing with AprilTag references.}\label{fig:pos}
\end{figure}

\begin{table}[tb]
\caption{Static endpoint error comparison. Full=Full resolution. Low=Low resolution. Adapt=Adaptive hyperparameters. Fixed=Fixed hyperparameter. Fix\_Agg=Fixed hyperparameter with more aggressive settings. 
Ours: Full/Adapt at 5 iterations evaluated over combined (Cmb), forward (Fwd), and reverse (Rev) directions. 
FP: FoundationPose.
FP*: FoundationPose results excluding method-specific failure cases.
We only report combined forward and backward results for the ablations and baselines due to space considerations as directional differences are minimal.}\label{tab:comparison}
\scriptsize
\setlength{\tabcolsep}{1.5pt}
\begin{tabular}{@{}lcccccccccccccc@{}}
\toprule
 & \multicolumn{3}{c}{\textbf{Ours}} & \multicolumn{7}{c}{\textbf{Ablation Study}} & \multicolumn{2}{c}{\textbf{Baseline}} \\
\cmidrule(lr){2-4}\cmidrule(lr){5-11}\cmidrule(lr){12-13}
 & \multicolumn{3}{c}{5 iter Full/Adapt} & \multicolumn{3}{c}{5 iter} & \multicolumn{4}{c}{10 iter} & \textbf{FP} & \textbf{FP*} \\
\cmidrule(lr){2-4}\cmidrule(lr){5-7}\cmidrule(lr){8-11}
\textbf{Metric} & Cmb & Fwd & Rev & \shortstack{Low\\Adapt} & \shortstack{Full\\Fixed} & \shortstack{Full\\Fix\_Agg} & \shortstack{Full\\Adapt} & \shortstack{Low\\Adapt} & \shortstack{Full\\Fixed} & \shortstack{Full\\Fix\_Agg} & & \\
\midrule
\multicolumn{13}{c}{\textbf{Translation (cm)}} \\
\midrule
Mean & 1.76 & 1.84 & 1.68 & 2.07 & 1.84 & 3.19 & 1.73 & 2.08 & 1.79 & 2.32 & 57.76 & 4.37 \\
Std & 0.69 & 0.68 & 0.70 & 0.78 & 0.76 & 1.72 & 0.67 & 0.75 & 0.74 & 1.01 & 112.32 & 1.49 \\
Min & 0.48 & 0.48 & 0.54 & 0.65 & 0.50 & 0.78 & 0.54 & 0.72 & 0.60 & 0.16 & 2.58 & 2.58 \\
Max & 3.23 & 3.06 & 3.23 & 3.70 & 3.55 & 7.36 & 3.13 & 3.67 & 3.10 & 4.61 & 329.92 & 6.73 \\
\midrule
\multicolumn{13}{c}{\textbf{Rotation (deg)}} \\
\midrule
Mean & 0.61 & 0.62 & 0.59 & 0.69 & 0.64 & 0.80 & 0.60 & 0.69 & 0.61 & 0.65 & 24.51 & 1.43 \\
Std & 0.23 & 0.27 & 0.18 & 0.25 & 0.22 & 0.43 & 0.23 & 0.25 & 0.22 & 0.26 & 48.45 & 0.57 \\
Min & 0.21 & 0.21 & 0.26 & 0.15 & 0.29 & 0.22 & 0.21 & 0.17 & 0.26 & 0.19 & 0.57 & 0.57 \\
Max & 1.28 & 1.28 & 1.06 & 1.39 & 1.25 & 2.15 & 1.29 & 1.35 & 1.34 & 1.19 & 136.60 & 2.46 \\
\botrule
\end{tabular}
\end{table}
Our method achieves sub-2~cm precision ($1.76 \pm 0.70$~cm combined) at 30+~Hz, with consistent bidirectional performance across forward ($1.84 \pm 0.68$~cm) and reverse ($1.68 \pm 0.70$~cm) trajectories.
The FoundationPose baseline exhibits significant degradation ($57.76 \pm 112.32$~cm) due to three additional failure sequences, in which the robot base was misclassified against surrounding structures in the depth maps. Excluding these failure cases, FoundationPose (marked as FP* in the results tables) yields $4.37 \pm 1.49$~cm, still $2.5\times$ less accurate than our method evaluated over all sequences.
\paragraph{Motion pose estimation against marker-based approach}\label{sec:posevsmarker}
\tabref{tab:position_error_comparison} compares pose estimation accuracy against the FoundationPose baseline with common marker-based pose estimation during robot in motion. 
Our method achieved near-1cm accuracy across 27,460 frames (forward and reverse combined): mean 3D error of $1.24$~cm with consistent bidirectional performance between forward ($1.22$~cm) and reverse ($1.26$~cm) directions, versus $27.40$~cm for the baseline. 
Excluding baseline outlier results (marked as FP* in the tables)
yielded an average 3D position error of $1.38$~cm, still 11\% worse than our method evaluated on all trajectories. 
FP* shows 44\% higher Y-axis error ($0.49$ vs $0.34$~cm) and 15\% higher Z-axis error ($1.06$ vs $0.92$~cm), with bias ratios of 81\% and 100\% respectively, indicating systematic offset rather than random variation. While FP* shows no frames exceeding 5~cm error, our method has 1.7\% above this threshold (max $14.46$~cm) due to render-segmentation disparity under rapid motion with insufficient gradient information. Nevertheless, our method achieves $<1cm$ accuracy in 54.7\% of frames versus 35.1\% for FP*, demonstrating that FP*'s lower variance reflects consistent bias rather than accuracy, as shown in \figref{fig:pos}

\begin{table}[tb]
\centering
\caption{Dynamic per-frame error comparison (cm). Full=Full resolution. Low=Low resolution. Adapt=Adaptive hyperparameters. Fixed=Fixed hyperparameter. Fix\_Agg=Fixed hyperparameter with more aggressive settings. 
Ours: Full/Adapt at 5 iterations evaluated of combined (Cmb), forward (Fwd), and reverse (Rev) directions. 
FP: FoundationPose.
FP*: FoundationPose results excluding method-specific failure cases.
We only report combined forward and backward results for the ablations and baselines due to space considerations as directional differences are minimal.}
\label{tab:position_error_comparison}
\scriptsize
\setlength{\tabcolsep}{1.2pt}
\begin{tabular}{@{}llcccccccccccccc@{}}
\toprule
& & \multicolumn{3}{c}{\textbf{Ours}} & \multicolumn{7}{c}{\textbf{Ablation Study}} & \multicolumn{2}{c}{\textbf{Baseline}} \\
\cmidrule(lr){3-5}\cmidrule(lr){6-12}\cmidrule(lr){13-14}
& & \multicolumn{3}{c}{5 iter Full/Adapt} & \multicolumn{3}{c}{5 iter} & \multicolumn{4}{c}{10 iter} & \textbf{FP} & \textbf{FP*} \\
\cmidrule(lr){3-5}\cmidrule(lr){6-8}\cmidrule(lr){9-12}
& & Cmb & Fwd & Rev & \shortstack{Low\\Adapt} & \shortstack{Full\\Fixed} & \shortstack{Full\\Fix\_Agg} & \shortstack{Full\\Adapt} & \shortstack{Low\\Adapt} & \shortstack{Full\\Fixed} & \shortstack{Full\\Fix\_Agg} & & \\
\midrule
\multicolumn{2}{l}{FPS} & 33 & 33 & 33 & 35 & 33 & 32 & 22 & 23 & 23 & 23 & -- & -- \\
\midrule
\multirow{2}{*}{X} & $\mu$ & 0.54 & 0.53 & 0.54 & 0.57 & 0.49 & 0.88 & 0.50 & 0.53 & 0.45 & 0.68 & 13.14 & 0.52 \\
 & $\sigma$ & 0.43 & 0.42 & 0.44 & 0.45 & 0.40 & 0.66 & 0.38 & 0.39 & 0.30 & 0.52 & 4.36 & 0.33 \\
\midrule
\multirow{2}{*}{Y} & $\mu$ & 0.34 & 0.34 & 0.34 & 0.40 & 0.29 & 0.76 & 0.29 & 0.34 & 0.25 & 0.50 & 7.24 & 0.49 \\
 & $\sigma$ & 0.38 & 0.38 & 0.38 & 0.40 & 0.27 & 0.54 & 0.30 & 0.31 & 0.22 & 0.39 & 2.15 & 0.29 \\
\midrule
\multirow{2}{*}{Z} & $\mu$ & 0.92 & 0.90 & 0.95 & 1.02 & 1.00 & 1.41 & 0.81 & 0.91 & 0.82 & 1.07 & 19.12 & 1.06 \\
 & $\sigma$ & 0.91 & 0.91 & 0.91 & 0.99 & 1.16 & 1.08 & 0.73 & 0.79 & 0.79 & 0.83 & 6.95 & 0.56 \\
\midrule
\multirow{2}{*}{3D} & $\mu$ & 1.24 & 1.22 & 1.26 & 1.37 & 1.25 & 2.06 & 1.11 & 1.23 & 1.05 & 1.52 & 27.40 & 1.38 \\
 & $\sigma$ & 0.95 & 0.95 & 0.95 & 1.01 & 1.17 & 1.02 & 0.75 & 0.80 & 0.78 & 0.81 & 6.93 & 0.57 \\
\midrule
\multicolumn{2}{l}{max} & 14.46 & 13.66 & 14.46 & 18.13 & 24.33 & 15.61 & 12.80 & 11.50 & 17.71 & 10.59 & 246.90 & 4.98 \\
\multicolumn{2}{l}{$<$1cm (\%)} & 54.7 & 55.5 & 53.9 & 46.8 & 57.1 & 11.7 & 58.8 & 49.8 & 63.1 & 27.5 & 28.5 & 35.1 \\
\multicolumn{2}{l}{$>$5cm (\%)} & 1.7 & 1.6 & 1.8 & 1.7 & 2.5 & 2.2 & 0.7 & 0.8 & 1.0 & 0.8 & 18.8 & 0.0 \\
\botrule
\end{tabular}
\end{table}

\figref{fig:pos} illustrates tracking performance during robot motion between positions 11 and 12. Our method achieves a mean 3D error of $1.10 \pm 0.74$~cm (median $0.83$~cm) across 643 valid frames (frames where the AprilTag was detected in both left and right images) versus $1.36 \pm 0.69$~cm (median $1.13$~cm) for the baseline, with error spikes during rapid motion transitions (frames 100--300). Our method exhibits minimal systematic bias (X: $-0.40$~cm, Y: $+0.05$~cm, Z: $-0.04$~cm), whereas the baseline shows pronounced directional biases (X: $+0.40$~cm, Y: $-0.58$~cm, Z: $+0.99$~cm), revealing systematic offset from ground truth, which is further agreed with the 2.5~cm error at the final frame for baseline comparing against 1~cm our our method.

\paragraph{Ablation study}
Full-resolution rendering improves static accuracy by 15.0\% ($1.76$ vs $2.07$~cm combined; $1.84$ vs $2.11$~cm forward, $1.68$ vs $2.03$~cm reverse) with 11.6\% rotation reduction ($0.61$\textdegree vs $0.69$\textdegree) at 2 FPS throughput cost. Dynamic evaluation shows 9.5\% lower error ($1.24$ vs $1.37$~cm) with 16.9\% more $<1cm$ error frames. Adaptive tuning yields 4.3\% static improvement ($1.76$ vs $1.84$~cm); dynamic accuracy is comparable ($1.24$ vs $1.25$~cm), but fixed hyperparameters exhibit 68.3\% higher maximum error and 23.2\% higher variance due to reduced motion responsiveness.

Aggressive fixed optimisation degrades accuracy by 66.1\% ($2.06$ vs $1.24$~cm combined; $2.05$ vs $1.22$~cm forward, $2.08$ vs $1.26$~cm reverse) with 78.6\% fewer $<1$~cm error frames due to premature convergence. Increasing to 10 iterations improves accuracy but reduces throughput to 22~FPS, below the 30+~FPS target, with marginal gain over adaptive mode in conservative fixed mode. The aggressive optimiser still suffers from premature convergence, as evidenced by the lowest maximum error and a low standard deviation, yielding 22\% worse accuracy than the adaptive settings at 5 iterations. Full-resolution adaptive at 5 iterations is optimal, achieving near 1~cm accuracy at 33~FPS.

\begin{figure}[tb]
\centering
\includegraphics[width=1\textwidth]{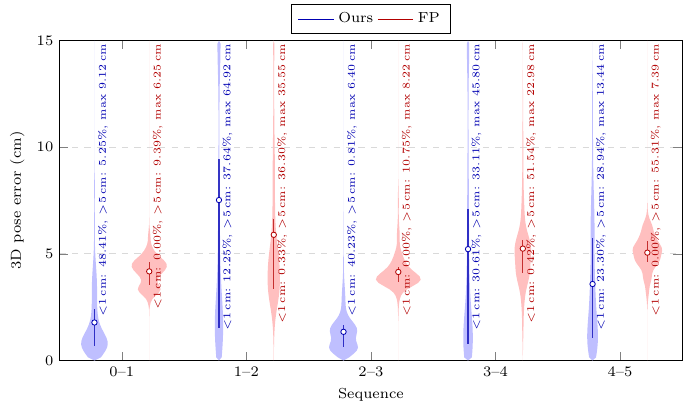}
\caption{%
Per-frame 3D pose error distributions (cm) of our method against the FoundationPose (FP) baseline under partial occlusion during robot motion, broken down by displacement sequence. Ours: Full/Adapt at 5 iterations evaluated at the forward sequence. Violins show the per-frame error distribution over all successfully detected frames in each sequence (Ours: 628/898/614/1199/781; FP: matched). Inner box: interquartile range; white dot: mean. Rotated labels report the percentage of frames below 1\,cm, the percentage above 5\,cm, and the maximum per-frame error. Per-frame errors are clipped to 15\,cm; frames with larger errors contribute mass at the ceiling.\label{fig:violin_occlude_all}}
\end{figure}

\paragraph{Occlusion study}
\figref{fig:violin_occlude_all} reports the per-frame 3D error distributions across all five displacement sequences. Under mild occlusion (Sequences 0\_1 and 2\_3), our model maintains performance close to the unoccluded results reported in \tabref{tab:position_error_comparison} (mean 3D error 1.24~cm, 54.1\% of frames below 1~cm, 1.17\% above 5~cm), with mean 3D errors of 1.78~cm and 1.35~cm, 48.41\% and 40.23\% of frames below 1~cm, and only 5.25\% and 0.81\% above 5~cm. FP, by contrast, degrades sharply from its unoccluded performance (1.38~cm mean, 35.1\% below 1~cm, 0\% above 5~cm) to mean errors of 4.18~cm and 4.15~cm, with almost no frames achieving sub-centimetre accuracy while 9.39\% and 10.75\% exceeding 5~cm. Under severe occlusion (Sequences 1\_2, 3\_4, and 4\_5), our model's distribution broadens substantially, with mean 3D errors of 7.52~cm, 5.22~cm, and 3.59~cm and per-frame maxima of 64.92~cm, 45.80~cm, and 13.44~cm, reflecting degraded segmentation under heavy occlusion that supplies inaccurate gradient information to the optimiser and yields less accurate pose estimates. FP yields mean errors of 5.89~cm, 5.25~cm, and 5.06~cm, and lower maxima of 35.55~cm, 22.98~cm, and 7.39~cm, consistent with the pattern observed in \tabref{tab:position_error_comparison}, with 36.30\%, 51.54\%, and 55.31\% of frames exceeding 5~cm. Despite higher maxima, our model still achieves sub-centimetre accuracy on 12.25\%, 30.61\%, and 23.30\% of frames, compared with 0--0.42\% for FP, and exceeds 5~cm on only 37.64\%, 33.11\%, and 28.94\% of frames. The systematic bias pattern observed in the unoccluded setting in \tabref{tab:position_error_comparison} persists here, with bias ratios of 82\% in X and 100\% in both Y and Z for the baseline.

\section{Discussion and conclusion}\label{sec5}
This work is the first to demonstrate the deployment of stereo-differentiable rendering for dynamic pose estimation scenarios, achieving online inference rates greater than the maximum camera frame rate of $30\,\text{fps}$ at 1080p, thus real-time performance, while maintaining 
near-1cm accuracy in unobstructed and 1.5 cm accuracy in mild occluded settings. We implemented parallelised CUDA stream for concurrent processing enabled accelerated runtime suitable for this online pose tracking scenario during robot displacement studies. This approach doubled the iteration rate compared to the original \roboreg pipeline while incorporating a more efficient optimisation algorithm.
Our results demonstrate that rendering-based pose estimation can achieve real-time performance through careful adaptation of the pipeline, without requiring fundamental algorithmic modifications, while preserving 1-cm level accuracy. This stands in contrast to existing differentiable rendering approaches: EasyHeC~\cite{Chen_2023} requires approximately 30~seconds per pose, and the comparison study in CtrNet~\cite{lu2023markerlesscameratorobotposeestimation} reports a loop rate of only 1~Hz, both of which are far below the 34~FPS achieved in this work.
Furthermore, compared to neural approaches such as CtRNet, our method provides superior interpretability through its white-box structure on direct dense correspondence, enabling direct pose estimation monitoring during operation.
Extensive evaluations against marker-based and markerless online tracking approaches confirmed applicability in dynamic scenarios and improvements over state of the art were found. In contrast to marker-based tracking, susceptible to motion blur, we found $13.4\%$ relatively improved tracking reliability. Compared to the markerless FoundationPose~\cite{10655554}, the proposed approach did not require depth, which is generally not available for draped surgical robots, and demonstrated on-par localisation. It consumed significantly less memory and achieved $6\times$ faster tracking; however, it relied on camera pose initialisation. 
Retaining \roboreg's platform-agnostic design, this method supports any robot with known CAD meshes.

Several limitations remain. Silhoutte-based differentiable rendering depends on segmentation accuracy, which directly affects pose estimation performance. Misclassifications, while mitigated by our robust loss, led to localisation degradation. This is corroborated by our results on the occlusion datasets, where the severe occluded segmentation mask increase the mean dynamic pose estimation error to 4.31~cm. While online tracking capabilities were demonstrated, we left the assumption of an initially known camera pose unaddressed. The proposed adaptive optimisation scheme helped mitigate some of the discrepancies inherent to gradient-based approaches, achieving both fast convergence and robust precision during runtime with complex motion, but could not compensate for all residual errors, particularly under rotation, also the effectiveness of tuning degradated under sustained rapid motion. However, processing 10~Hz sequences at 30~Hz yields only one-third as many unique frames as native 30~Hz acquisition, reducing temporal resolution and potentially degrading inter-frame initialisation under fast motion. Higher-frame-rate data collection will address this limitation in future work. Although the \roboreg library already supports multi-robot setups, for both registration and segmentation, investigating deployment was beyond the scope of this work, and will be addressed in future research. The original \roboreg demonstrated occlusion and draping handling capability. 
Our occlusion study confirms that it is indeed robust under mild occlusion but nonetheless exhibits substantial degradation under severe occlusion in dynamic settings, motivating future work to improve performance in this regime. Future validation on a draped robot in dynamic scenario is also necessary to broaden applicability to practical scenarios.
The current study is limited to a single KUKA LBR Med 7 platform with data collected in a controlled laboratory environment. Future work should extend the evaluation to additional platforms, potentially the CMR Versius system or xArm7, using both realistic and synthetic datasets to produce more diverse and robust evaluations closer to actual deployment scenarios, thereby enabling further methodological comparison with other state-of-the-art methods. Future data collection should also include more clinically representative settings, such as video captured in simulated surgical suites.
Lastly, in our deployment scenario, the robot kinematics are fixed. Additional experiments, including robot articulation changes during movement, can be conducted in the future to evaluate the model's robustness to joint variations during deployment.

\section*{Declarations}

\paragraph{Author contributions}
Y.H., T.V., M.H. conceptualised runtime improvements. Y.H. contributed adaptive optimisation and performed evaluations. M.H., Y.H. collected benchmarking data. Y.H., M.H., C.B., T.V. authored and revised the manuscript. M.H. contributed the stereo differentiable algorithm, core software components, the experimental setup and associated robot drivers. M.H. collected ground-truth segmentation data, developed and trained all deep learning models.

\paragraph{Interests}
TV is co-founder and shareholder of Hypervision Surgical Ltd, London, UK.
The authors have no other relevant interests to declare.

\paragraph{Funding}
This work was supported by the EPSRC CDT in Advanced Engineering for Personalised Surgery \& Intervention [EP\/Y035364\/1].
This project received funding by the National Institute for Health and Care Research (NIHR) under its Invention for Innovation (i4i) Programme [NIHR202114].
The views expressed are those of the author(s) and not necessarily those of the NIHR or the Department of Health and Social Care.
For the purpose of open access, the authors have applied a CC BY public copyright license to any Author Accepted Manuscript version arising from this submission.

\bibliography{sn-bibliography}

\end{document}